\documentclass[runningheads,a4paper]{llncs}

\usepackage{amssymb}
\usepackage{graphicx}

\usepackage{url}

\urldef{\mailsc}\path|noswa023@uottawa.ca|    

\newcommand{\keywords}[1]{\par\addvspace\baselineskip
\noindent\keywordname\enspace\ignorespaces#1}

\begin{document}

\mainmatter  

\title{SemEval-2019 (OffensEval): Identifying and Categorizing Offensive Language in Social Media}

%
%
\author{Nikhil Oswal}
%


\institute{School of Electrical Engineering and Computer Science (EECS),\\ University of Ottawa,\\
Ottawa, Canada\\
\mailsc
}
%
%

\titlerunning{Identifying and Categorizing Offensive Language in Social Media}

\maketitle

\begin{abstract} 

Offensive language is pervasive in social media. Individuals frequently take advantage of the perceived anonymity of computer-mediated communication, using this to engage in behavior that many of them would not consider in real life. The automatic identification of offensive content online is an important task that has gained more attention in recent years. This task can be modeled as a supervised classification problem in which systems are trained using a dataset containing posts that are annotated with respect to the presence of some form(s) of abusive or offensive content. The objective of this study is to provide a description of a classification system built for SemEval-2019 Task 6: OffensEval. This system classifies a tweet as either offensive or not offensive (Sub-task A) and further classifies offensive tweets into categories (Sub-tasks B \& C). We trained machine learning and deep learning models along with data preprocessing and sampling techniques to come up with the best results. Models discussed include Naive Bayes, SVM, Logistic Regression, Random Forest and LSTM.
\end{abstract}

\section{Introduction}

With so many social media platforms existing on the internet
today, the amount of content created online on a daily basis is
growing at an ever-increasing rate as more of our world becomes
digitized \cite{1}. Social media websites have been focusing on
encouraging people to participate in content generation, but have
paid less attention to the content itself.

Identifying and eliminating ‘offensive / toxic’ content has become a
problem for any major website today. By the time a user reports
any harmful content and any action is taken by the website, the
content could have done a ton of damage already. Social media
websites such as YouTube have recently started taking serious
actions in order to remove problematic content from their
websites \cite{2}\cite{3}. Facebook rolled out a ‘protective detection’
system designed to flag posts that come from people threatening
suicide or self-harm and posts that are aggressive towards others \cite{4}.

Our study focuses on the identification and categorization of such offensive language in social media. It
focuses on three subtasks namely offensive language detection, categorization of offensive language and offensive language target identification.
Sub Task A aims to detect text as offensive (OFF)
or not offensive (NOT). Sub Task B aims to categorize the offensive type as targeted text (TIN)
or untargeted text (UNT). Sub Task C focuses on
the identification of target as individual (IND), group
(GRP) or others (OTH). 

The remainder of this paper is organized as follows. First, we describe the task and the dataset under consideration in Section 2. The adopted methods and techniques are presented in Section 3, while the experiments are discussed in Section 4. Section 5 talks about the results obtained by other participants. Finally, closing conclusion and future work are discussed in section 6.

\section{Case Study}

\subsection{Task Description}
In this study, the task under consideration is divided into three sub-tasks as follows.\hfill \break
\hfill \break
\textbf{Sub-task A - Offensive language identification:}
In this sub-task, we are interested in the identification
of offensive posts and posts containing any form of
(untargeted) profanity. In this sub-task, there are 2
categories in which the tweet could be classified -

\textbf{Not Offensive} - This post does not contain offense or
profanity.

\textbf{Offensive}
 - This post contains offensive language or
a targeted (veiled or direct) offense.
To sum up, this category includes insults, threats, and
posts containing profane language and swear words. \hfill \break
\hfill \break
\textbf{Sub-task B - Automatic categorization of offense
types:}
In this sub-task, we are interested in categorizing
offenses. Tweets are labeled from one of the
following categories -

\textbf{Targeted Insult} - A post containing an insult or a
threat to an individual, group, or others.

\textbf{Untargeted}
 - A post containing non-targeted
profanity and swearing.

\noindent Posts containing general profanity are not targeted
but they contain non-acceptable language. On the
other hand, insults and threats are targeted at an
individual or group.  \hfill \break
\hfill \break
\textbf{Sub-task C - Offense target identification:}
Finally, in sub-task C we are interested in the target
of offenses. Only posts that are either insults or
threats are included in this sub-task. The three
categories included in sub-task C are the following - 

\textbf{Individual} - The target of the offensive post is an
individual: a famous person, named individual or an
unnamed person interacting in the conversation.

\textbf{Group}
 - The target of the offensive post is a group of
people considered as a unity due to the same
ethnicity, gender or sexual orientation, political
affiliation, religious belief, or something else.

\textbf{Other}
 - The target of the offensive post does not
belong to any of the previous two categories (e.g. an
organization, a situation, an event, or an issue).

\subsection{Dataset}

In our study, we have used the OLID dataset given by OffensEval - SemEval2019 shared task. The dataset is
given in .tsv file format with columns namely,
ID, INSTANCE, SUBA, SUBB, SUBC where ID
represents the identification number for the tweet,
INSTANCE represents the tweets, SUBA consists
of the labels namely Offensive (OFF) and Not
Offensive (NOT), SUBB consists of the labels
namely Targeted Insult and Threats (TIN) and Untargeted (UNT) and SUBC consists of the labels
namely Individual (IND), Group (GRP) and Other
(OTH). The dataset comprised of 14,100 annotated tweets divided into a training partition of 13,240 tweets and
a testing partition of 860 tweets. 

\begin{figure}[hbt!]
\centering
\frame{\includegraphics[width=0.5\textwidth]{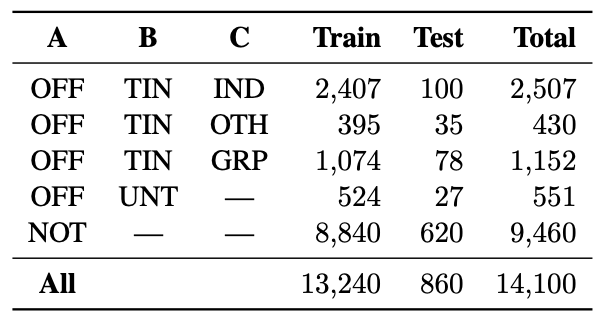}}
    \caption{Dataset} 
\end{figure}

\begin{figure}[hbt!]
\centering
\frame{\includegraphics[width=0.9\textwidth]{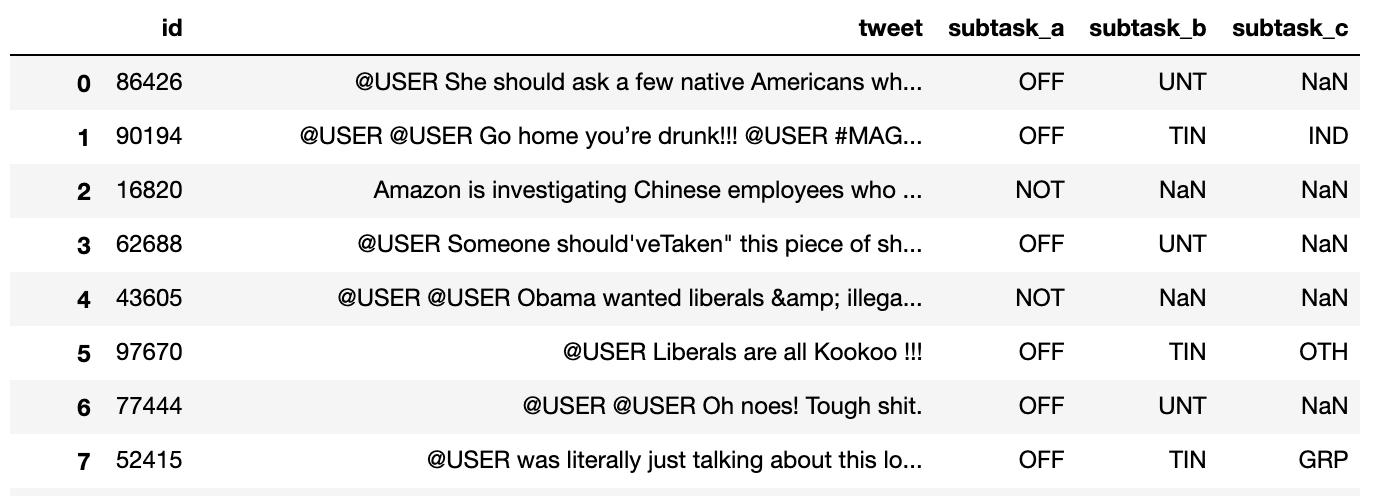}}
    \caption{Sample Instances} 
\end{figure}

\section{Methodology}
The overall architecture includes five major components: Data Exploration and Analysis, Data Pre-processing, Feature Extraction, Model Implementation, and Model Evaluation.
\subsection{Data Exploration and Analysis}
Exploratory Data Analysis is valuable to machine learning problems since it allows to get closer to the certainty that the future results will be valid, correctly interpreted, and applicable to the desired business contexts \cite{5}. We performed EDA using pandas and matplotlib.

As mentioned before, the dataset comprises of 14,100 tweets partitioned into 13,240 as training data and 860 as test data. Upon further exploring the data, we can see that we have 67\% of tweets as Not Offensive and 33\% tweets as Offensive. Figure 3, 4 and 5 shows the data distribution of each of the subtasks. We can say that that dataset for the later too tasks are highly imbalanced.

\begin{figure}[hbt!]
\centering
\frame{\includegraphics[width=0.4\textwidth]{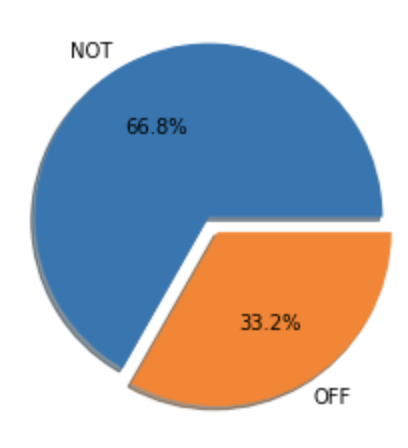}}
    \caption{Task A Dataset Distribution} 
\end{figure}
\begin{figure}[hbt!]
\centering
\frame{\includegraphics[width=0.4\textwidth]{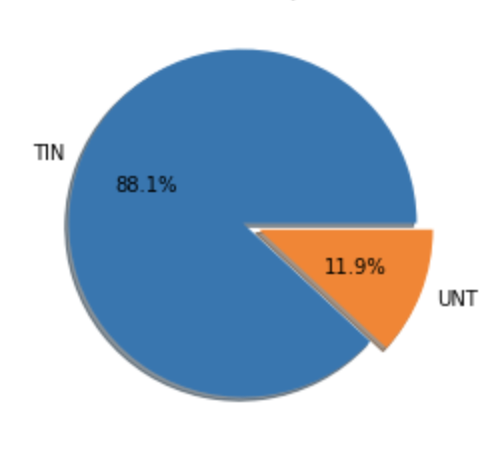}}
    \caption{Task B Dataset Distribution} 
\end{figure}
\begin{figure}[hbt!]
\centering
\frame{\includegraphics[width=0.4\textwidth]{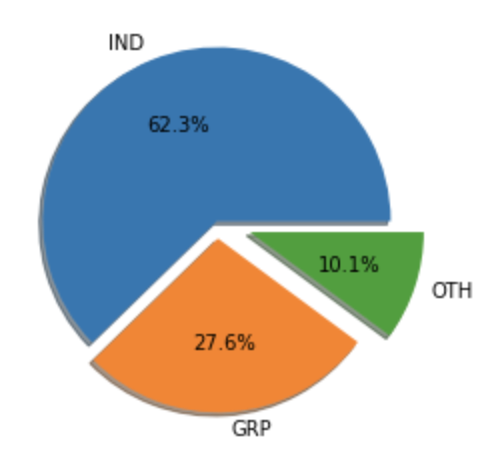}}
    \caption{Task C Dataset Distribution} 
\end{figure}

To analyze the word frequency distribution in the offensive tweets, we first need to clean the data and remove words
that are not meaningful. We applied some data cleaning
techniques such as - tokenizing sentence to words, converting
words to lowercase, removing punctuations, removing stop
words. When we
plot the Word frequency distribution chart (Figure. 6), we
observed that the most frequently used words in the offensive tweets were – liberals, control, people, antifa, Trump, etc.

\begin{figure}[hbt!]
\centering
\frame{\includegraphics[width=0.9\textwidth]{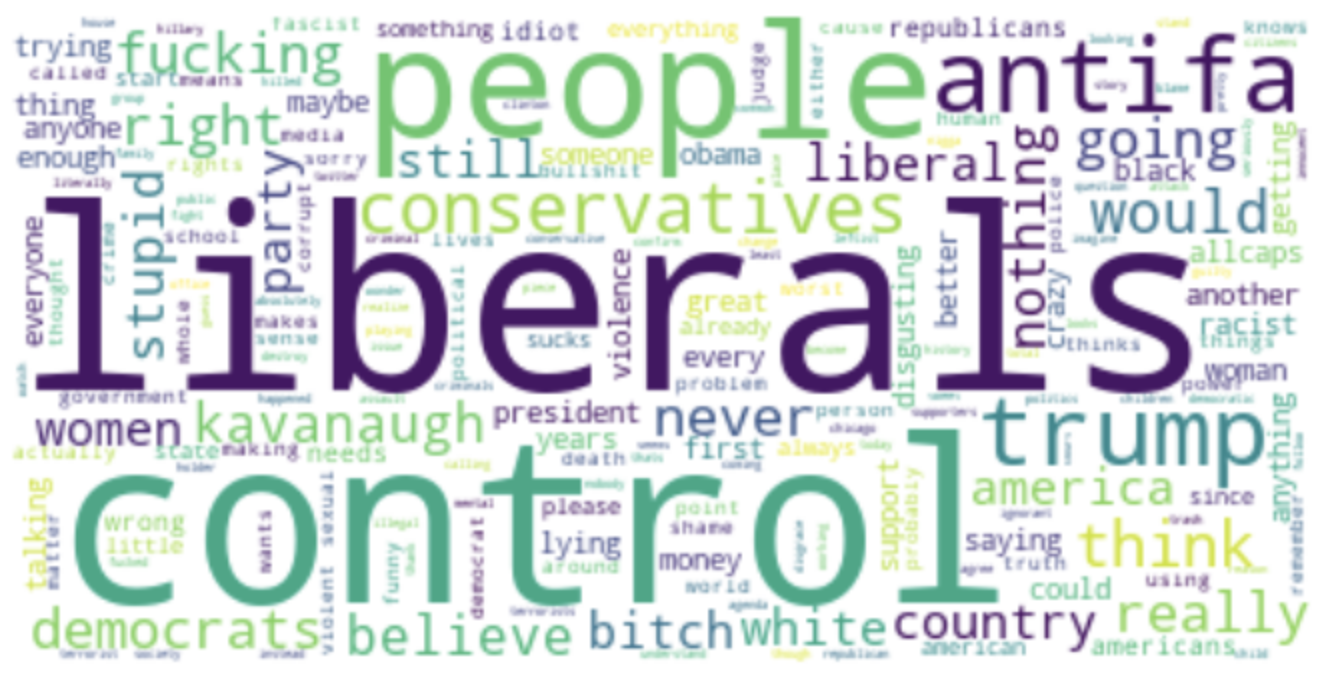}}
    \caption{Word Cloud of the most frequent words} 
\end{figure}

\begin{figure}[hbt!]
\centering
\frame{\includegraphics[width=\columnwidth]{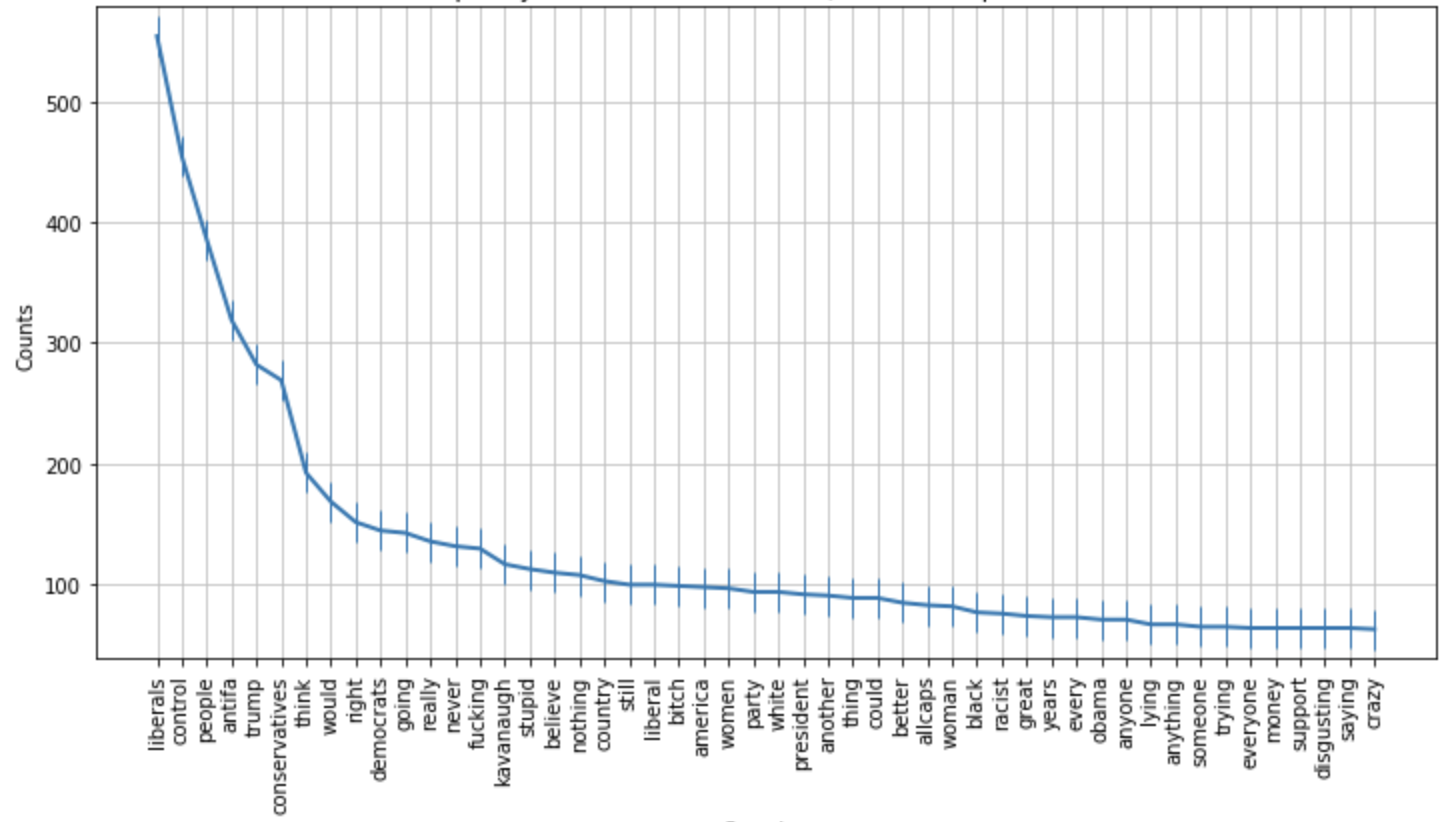}}
    \caption{Frequency Distribution of words} 
\end{figure}

\subsection{Data Preprocessing}
The dataset provided contains raw twitter data. By its nature, this data contains a lot of hard to interpret features, so it would have been unwise to apply vectorization or word embeddings directly to the
unprocessed tweets. Some of these features include informal language, grammatical mistakes, emojis
and special characters. We therefore undertook to perform extensive pre-processing work on the raw
data before applying our models on it.

\subsubsection{Twitter Specific:} As we are dealing with natural language data, we were able to use regex to
apply powerful transformations that remove every \#, transforms every @USER into the tag $<user>$. Also as many hashtags contain words of the form ”ILove...”, we apply some regular expression
(regex) formulas in order to split those in ``I Love ...”.

\subsubsection{Emojis:} Emojis are widely present across the dataset. Most of the time they represent a feeling
or contextually relevant information. We have manually created a dictionary of the emojis which we considered to be most
important based on a qualitative analysis of the data, and replaced the emojis identified as such with
their associated transcriptions.

\subsubsection{Special Characters:} We removed most of the special characters except dots,
commas, question marks, and exclamation marks. These are frequent punctuation characters, which
are also included in the embedding. \hfill \break

\noindent Apart from this, we removed white spaces, numbers and repeated tokens. We also handled tweets which were in full caps by appending them with $<allcaps>$ tags.

\subsubsection{Tokensiation:} Once the cleaning was done, the text of each tweet is split at blank spaces. We
added blank spaces between punctuation and other words where necessary to ensure the tokenization
separated them.

\subsection{Feature Extraction}

After preprocessing and cleaning our textual data, we next transform our dataset to numerical data as most models only work
with numerical features. We used below techniques to perform feature extraction.

\subsubsection{TF-IDF / Count} are one of the simplest type of
vectorization techniques. \textbf{Count} first builds the vocabulary dictionary where
keys are the words available in the corpus and value
is the index of the word in a vector. The vector’s size
is the number of unique words in the corpus where 
each index is a word mapped by the vocabulary
dictionary and its value is the number of occurrence
of this word in the corresponding sentence. \textbf{TFIDF} stands for Term Frequency – Inverse
Document Frequency, starts just like count but
doesn’t just replace a word with its count, it replaces
a word according to the following formula.
\begin{center}
    $w_{i,j} = tf_{i,j} * log\frac{N}{df_i}$

\end{center}

\noindent Inverse Document Frequency (IDF) is defined as
logarithm of ratio of total samples available in the
corpus and number of samples containing a unique
word. TF IDF formula gives the relative importance
of a word in a corpus.

\subsubsection{Word Embedding} is one of the most popular
representation of document vocabulary. It is capable
of capturing context of a word in a document,
semantic and syntactic similarity, relation with other
words. Word embedding in basically a vector
representation of a word in a corpus, there are many
available models to work with, the following 2
variations are used.

\textbf{Glove:} GloVe stands for global vectors for word
representation \cite{6}. We used GloVe embedding’s
which is based on factorizing a matrix of word co-occurrence statistics. We experimented on all
dimensions of glove vectors and decided to work with
200-dimensional GloVe vectors as they provided the
best results for our task

\textbf{Keras Embedding Layer:} Keras offers an Embedding
layer that can be used for neural networks on text data
\cite{7}. The Embedding layer is defined as the first hidden
layer of a network with three arguments – input
dimension, output dimension and input length.

\subsection{Model}
\subsubsection{Naive Bayes} is based on Bayes’ theorem
with the naïve assumption of independence between
each pair of features. If we need to classify the vector X
= x1…xn into m classes, C1…Cm. we need to find the
probability of each class given X. Then we can assign X
the label of the class with highest probability. The
probability is calculated using Bayes’ theorem which is
defined as:

\begin{center}
    $P(C_i | X) = \frac{P(C_i | X) P(C_i)}{ P(X)}$
\end{center}

We used the Naïve Bayes classifier for multinomial
models provided by sklearn [8] with default
parameters. We chose the multinomial Naïve Bayes
classifier because is appropriate for text classification. 

\subsubsection{Support Vector Machine} works by finding and
constructing a hyperplane in N-dimensional space that
separates the points between two classes, N being the
number of features. The hyperplane is determined by
finding a plane that has the maximum margin which is
the distance between the data point of two classes.
Points that fall on the side of the hyperplane can be
attributed to different classes. We used the Support
Vector Classifier provided by sklearn \cite{svm} for training
and testing. The kernel type to be used in the algorithm
is `linear’ (x, x’). The degree of the polynomial kernel
function is chosen as 3. The gamma parameter is set to
`auto’ which uses 1/ n features.

\subsubsection{Random Forest}is a supervised ensemble learning algorithm. ‘Ensemble’ means that it takes a bunch of ‘weak learners’ and have them work together to form one strong predictor \cite{randomforest}. Here, we have a collection of decision trees, known as “Forest”. To classify a new object based on attributes, each tree gives a classification and we say the tree “votes” for that class. The forest chooses the classification having the most votes (over all the trees in the forest). 
\subsubsection{LSTM} is a feed
forward neural network. Vanilla RNN fail to understand
the context behind an input. They are not able to recall
some text that they saw long back to make predictions
in the present. LSTM are able to choose ehat
information should be remembered or which should be
forgotten. They make use of a forget gate, input gate
and output gate to do so. We have used LSTM implementation by keras \cite{lstm}.

Embeddings - Keras Embedding Layer, GloVe

Activation Function – Relu for the middle dense layer,
Sigmoid for the last dense layer.

Loss Function – Binary cross entropy.

Optimizer – Adam

Figure 7 depicts the architecture of our model.

\begin{figure}[hbt!]
\centering
\frame{\includegraphics[width=\columnwidth]{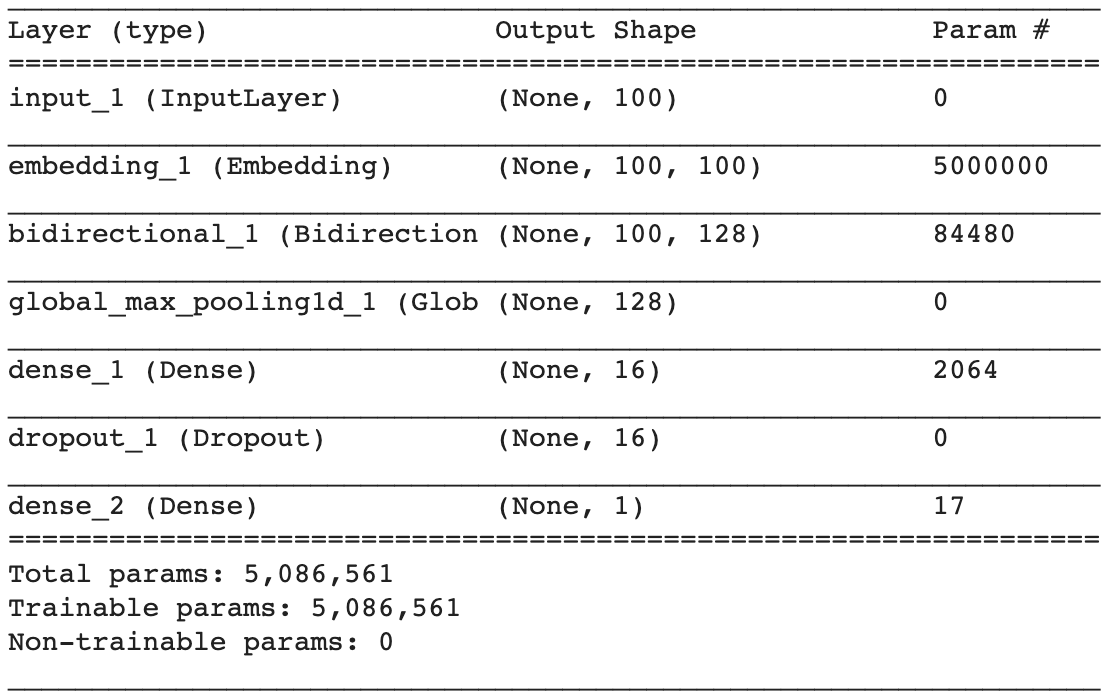}}
    \caption{Architecture of the Bidirectional LSTM model.} 
\end{figure}

\subsection{Evaluation}

\noindent\textbf{Accuracy:} The ratio of correct predictions over the total
predictions.

\begin{center}
    $Accuracy = \frac{TP +TN}{TP + TN + FP + FN}$
\end{center}
\noindent\textbf{Precision:} The number of true positives divided by all
positive predictions. It is a measure of a classifier’s exactness. It
tells us how often the classifier is correct when it predicts
positive. Low precision means that there is a high number of
false positives.

\begin{center}
   $ Precision = \frac{TP}{TP + FP}$
\end{center}

\noindent \textbf{Recall:} The number of true positives divided by the number
of positive values in the test data. It is also known as Sensitivity
or the True Positive Rate. It is a measure of a classifier’s
completeness. It tells us how often the classifier is correct for all
positive instances. Low recall means that there is a high number
of false negatives. 

\begin{center}
    $Recall = \frac{TP}{TP + FN}$
\end{center}

\noindent \textbf{F1-Score:} It is the harmonic mean of precision and recall.

\begin{center}
    $F1 Score = 2 *  \frac{Precision * Recall}{Precision + Recall}$
\end{center}

\noindent \textbf{Confusion Matrix:} It is a table that is used to describe the
performance of a classifier on the test data for which the true
values are known (Table 1). 

\begin{table}[]
\caption{Confusion Matrix}

\centering
\begin{tabular}{|l|l|l|}
\hline
 & \textit{\textbf{Predicted YES}} & \textit{\textbf{Predicted NO}} \\ \hline
\textit{\textbf{Actual YES}} & True Positives (TP) & False Negatives (FN) \\ \hline
\textit{\textbf{Actual NO}} & False Postives (FP) & True Negatives (TP) \\ \hline
\end{tabular}
\label{tab:my-table}
\end{table}

\section{Experiments}
For all the experiments and development of classifiers, we used Python 3 and Google colab's Jupyter Notebook. We used libraries such as Sckit Learn, Matplotlib, Seaborn, Pandas, Numpy and Imblearn. 

We carried experiments with different input data; one with the original dataset, then with the undersampled dataset and last one with the oversampled dataset. We splitted out dataset in ratio of 80:20 for training and testing purpose. Below we report some of the best results obtained.

\subsection{Subtask - A}
\begin{table}[]
\caption{Results for sub task A}

\centering
\begin{tabular}{|c|c|c|c|c|}
\hline
\textbf{Model} & \textbf{Data} & \textbf{Feature Extraction} & \textbf{Accuracy} & \textbf{F Score} \\ \hline
\textbf{Naïve Bayes} & Original Dataset & Count Vectorizer & 74.74 & 0.69 \\ \hline
\textbf{SVM} & Original Dataset & TF-ID & 77.49 & 0.70 \\ \hline
\textbf{SVM} & Original Dataset & Count Vectorizer & 74.13 & 0.69 \\ \hline
\textbf{SVM} & Oversampled & TF-ID & 75.10 & 0.71 \\ \hline
\textbf{Random Forest} & Undersampled & Count Vectorizer & 72.59 & 0.69 \\ \hline
\textbf{Logistic Regressiom} & Original Dataset & TF-ID & 75.89 & 0.67 \\ \hline
\textbf{Logistic Regressiom} & Original Dataset & Count Vectorizer & 76.04 & 0.70 \\ \hline
\textbf{LSTM} & Original Dataset & Keras EL & 72.29 & 0.68 \\ \hline
\textbf{LSTM} & Original Dataset & Glove &77.89  & 0.72 \\ \hline
\end{tabular}
\label{tab:my-table}
\end{table}

Based on Accuracy and F Score, we can say that LSTM performed the best.
			
\subsection{Subtask - B}

\begin{table}[]
\caption{Results for sub task B}

\centering
\begin{tabular}{|c|c|c|c|c|}
\hline
\textbf{Model} & \textbf{Data} & \textbf{Feature Extraction} & \textbf{Accuracy} & \textbf{F Score} \\ \hline
\textbf{Naïve Bayes} & Original Dataset & Count Vectorizer & 86.9 & 0.49 \\ \hline
\textbf{Naïve Bayes} & Original Dataset & TF-IDF & 88.09 & 0.46 \\ \hline
\textbf{SVM} & Original Dataset & TF-IDF & 83.9 & 0.51 \\ \hline
\textbf{SVM} & Oversampled & TF-ID & 83.45 & 0.58 \\ \hline
\textbf{Random Forest} & Oversampled & TF-ID & 84.72 & 0.62 \\ \hline
\textbf{Logistic Regressiom} & Original Dataset & Count Vectorizer & 88.18 & 0.53 \\ \hline
\textbf{LSTM} & Original Dataset & GloVe & 88.09 & 0.46 \\ \hline
\end{tabular}
\label{tab:my-table}
\end{table}
Based on Accuracy and F Score, we can say that Random Forest performed the best.

\subsection{Subtask - C}

\begin{table}[!hbt]
\caption{Results for sub task C}
\centering
\begin{tabular}{|c|c|c|c|c|}
\hline
\textbf{Model} & \textbf{Data} & \textbf{Feature Extraction} & \textbf{Accuracy} & \textbf{F Score} \\ \hline
\textbf{Naïve Bayes} & Original Dataset & Count Vectorizer & 69.83 & 0.44 \\ \hline
\textbf{SVM} & Oversampled Dataset & TF-IDF & 69.12 & 0.51 \\ \hline
\textbf{Logistic Regression} & Oversampled Dataset & Count Vectorizer & 60.12 & 0.48 \\ \hline
\textbf{LSTM} & Original Dataset & Keras EL & 62.26 & 0.26 \\ \hline
\textbf{LSTM} & Original Dataset & Glove & 70.96 &  0.46  \\ \hline
\end{tabular}

\label{tab:my-table}
\end{table}

Based on Accuracy and F Score, we can say that SVM performed the best.

\section{Results}

\begin{table}[]
\caption{Best Results}

\centering
\begin{tabular}{|c|c|c|c|}
\hline
\textit{\textbf{Subtask}} & \textit{\textbf{Model}} & \textit{\textbf{Accuracy}} & \textit{\textbf{F1 Score (macro)}} \\ \hline
\textit{\textbf{Task A}} & LSTM & 77.89 & 0.72 \\ \hline
\textit{\textbf{Task B}} & Random Forest & 84.72 & 0.62 \\ \hline
\textit{\textbf{Task C}} & SVM & 69.12 & 0.51 \\ \hline
\end{tabular}
\label{tab:my-table}
\end{table}
From the experiments above, it is evident that the Deep Learning model (LSTM) outperformed than our Machine Learning classifiers for subtask A. However, the same is not true for the other two tasks. For the subtask B and C, LSTM did not perform well in comparison to SVM. This might be because the number of instances for task B and C are too low. Below are the confusion matrix for the best models for each of the tasks.

\begin{figure}[hbt!]
\centering
\frame{\includegraphics[width=0.5\textwidth]{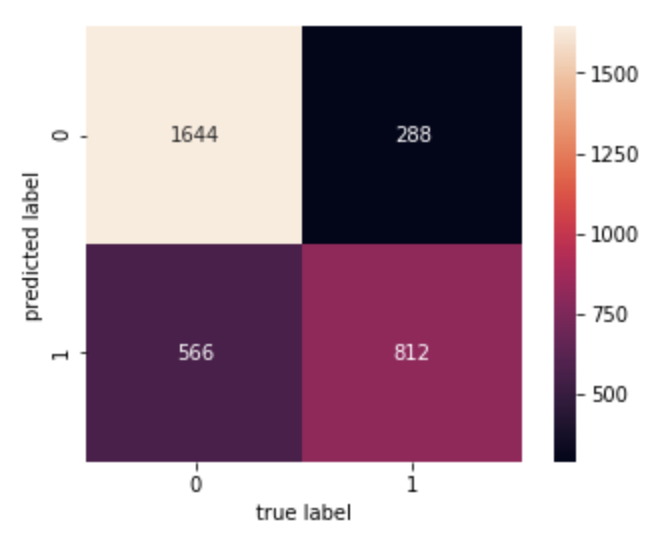}}
    \caption{Confustion Matrix - LSTM, Task - A} 
\end{figure}

\begin{figure}[hbt!]
\centering
\frame{\includegraphics[width=0.5\textwidth]{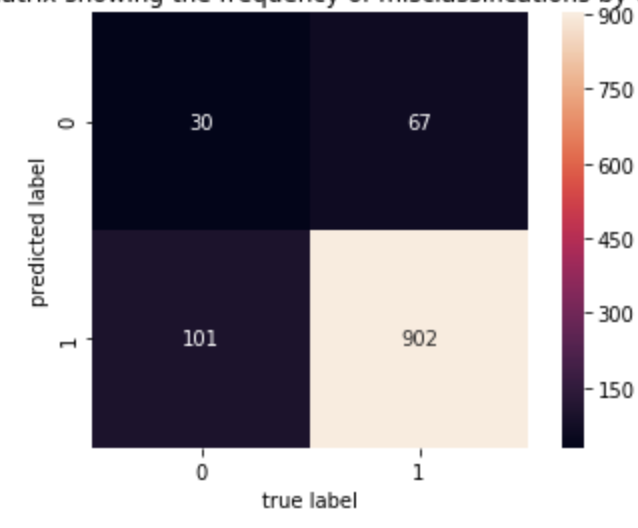}}
    \caption{Confustion Matrix - Naive Bayes, Task - B} 
\end{figure}

\begin{figure}[hbt!]
\centering
\frame{\includegraphics[width=0.5\textwidth]{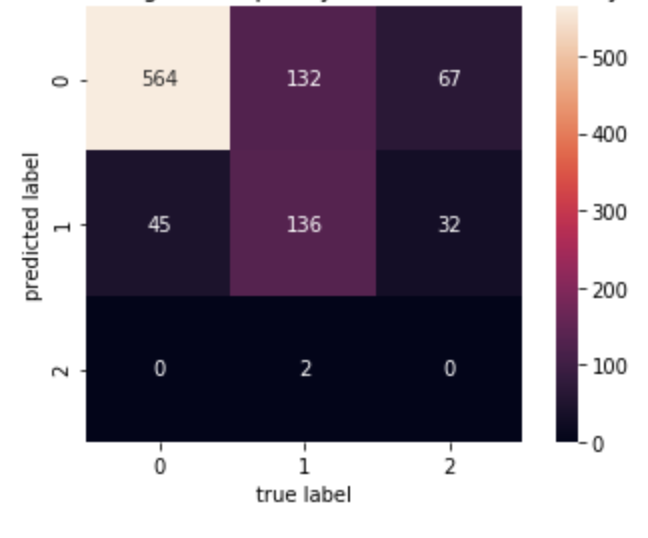}}
    \caption{Confustion Matrix - SVM, Task - C} 
\end{figure}

\subsubsection{Other Participants Results:}
The task had nearly 800 teams and 115 of
them submitted their results. The models used in the task submissions ranged
from traditional machine learning, e.g., SVM and
logistic regression, to deep learning, e.g., CNN,
RNN, BiLSTM, including attention mechanism,
to state-of-the-art deep learning models such as
ELMo \cite{peters} and BERT \cite{devlin}. Table 6. depicts the results of the top performing teams.

\begin{table}[]
\caption{Results of other participants}

\centering
\begin{tabular}{|c|c|c|c|c|c|}
\hline
\multicolumn{2}{|c|}{\textbf{Task A}} & \multicolumn{2}{c|}{\textbf{Task B}} & \multicolumn{2}{c|}{\textbf{Task C}} \\ \hline
\textbf{Team Ranks} & F1 & Team Ranks & F1 & Team Ranks & F1 \\ \hline
\textbf{1} & 0.829 & 1 & 0.755 & 1 & 0.66 \\ \hline
\textbf{2} & 0.815 & 2 & 0.739 & 2 & 0.628 \\ \hline
\textbf{3} & 0.814 & 3 & 0.719 & 3 & 0.626 \\ \hline
\textbf{4} & 0.808 & 4 & 0.716 & 4 & 0.621 \\ \hline
\textbf{5} & 0.807 & 5 & 0.708 & 5 & 0.613 \\ \hline
\textbf{6} & 0.806 & 6 & 0.706 & 6 & 0.613 \\ \hline
\textbf{7} & 0.804 & 7 & 0.7 & 7 & 0.591 \\ \hline
\textbf{8} & 0.803 & 8 & 0.695 & 8 & 0.588 \\ \hline
\textbf{9} & 0.802 & 9 & 0.692 & 9 & 0.587 \\ \hline
\textbf{10} & 0.8 & 10 & 0.69 & 10 & 0.586 \\ \hline
\end{tabular}
\label{tab:my-table}
\end{table}

\section{Conclusion and Future Work}
We were able to analyze the ‘OLID’ dataset to classify the tweet as offensive or
not offensive and further categorize them. We processed the dataset using various word
embedding models such as GloVe, Count Vectorizer and TF-IDF. We also tried oversampling technique using SMOTE although there was not much improvement in the performance. We trained various models such as SVM,
Naïve Bayes, Logistic Regression, Random Forest and LSTM. We observed that the deep model performed the best for task A and machine learning models performed well for the other tasks. The best results were obtained using LSTM (for task A), Random Forest (for task B) and SVM (for task C).

The evaluation results for this competition have shown that the best systems used ensembles
and state-of-the-art deep learning models such as
BERT \cite{results}. In the future work, the dataset could be trained using this models and its performance could be compared with our classifiers. The organizers of the competition also plans to increase the size of
the OLID dataset, while addressing issues such
as class imbalance and the small size for the test
partition, particularly for sub-tasks B and C. We plan to check the performance of our classifiers with the new dataset. With the additional data, we can certainly hope that Deep Learning model will perform better for the sub task B and C. The code and other resources is available at \url{https://github.com/nikhoswal/OffensEval-Task6}


\end{document}